\documentclass[conference]{IEEEtran}
\IEEEoverridecommandlockouts
\usepackage{cite}
\usepackage{amsmath,amssymb,amsfonts}
\usepackage{algorithmic}
\usepackage{graphicx}
\usepackage{textcomp}
\usepackage{xcolor}
\def\BibTeX{{\rm B\kern-.05em{\sc i\kern-.025em b}\kern-.08em
   T\kern-.1667em\lower.7ex\hbox{E}\kern-.125emX}}

\usepackage{booktabs}
\usepackage{subcaption}

\usepackage{bbm}

\begin{document}

\title{ConstraintMatch for Semi-constrained Clustering}

\author{

\IEEEauthorblockN{
Jann Goschenhofer\IEEEauthorrefmark{1}\IEEEauthorrefmark{2}\IEEEauthorrefmark{3},
Bernd Bischl\IEEEauthorrefmark{1}\IEEEauthorrefmark{2}\IEEEauthorrefmark{3}, 
Zsolt Kira\IEEEauthorrefmark{4}
}
\IEEEauthorblockA{
LMU Munich\IEEEauthorrefmark{1}, Fraunhofer Institute for Integrated Circuits\IEEEauthorrefmark{2}\\
Munich Center for Machine Learning\IEEEauthorrefmark{3}, Georgia Institute of Technology\IEEEauthorrefmark{4}
}

}

\maketitle

\begin{abstract}
\looseness=-1 Constrained clustering allows the training of classification models using pairwise constraints only, which are weak and relatively easy to mine, while still yielding full-supervision-level model performance.
While they perform well even in the absence of the true underlying class labels, constrained clustering models still require large amounts of binary constraint annotations for training.
In this paper, we propose a semi-supervised context whereby a large amount of \textit{unconstrained} data is available alongside a smaller set of constraints, and propose \textit{ConstraintMatch} to leverage such unconstrained data.
While a great deal of progress has been made in semi-supervised learning using full labels, there are a number of challenges that prevent a naive application of the resulting methods in the constraint-based label setting.
Therefore, we reason about and analyze these challenges, specifically 1) proposing a \textit{pseudo-constraining} mechanism to overcome the confirmation bias, a major weakness of pseudo-labeling, 2) developing new methods for pseudo-labeling towards the selection of \textit{informative} unconstrained samples, 3) showing that this also allows the use of pairwise loss functions for the initial and auxiliary losses which facilitates semi-constrained model training.
In extensive experiments, we demonstrate the effectiveness of ConstraintMatch over relevant baselines in both the regular clustering and overclustering scenarios on five challenging benchmarks and provide analyses of its several components.
\end{abstract}

%
%

\section{Introduction}
Manual annotation of class labels is a tedious and labor-intensive task that can constitute a significant obstacle in applications, particularly in situations where the annotator has to select from a large number of potential class labels or where the annotation is ambiguous due to the task complexity. 
Additionally, supervised classification models require knowledge of the total number of classes present in the respective application, i.e. the cardinality of the label space.
\textit{Constrained clustering} offers a remedy for this as model training in this weakly supervised regime requires only weak, pairwise constraint relations (i.e. similar/dissimilar) which incur less annotation effort compared to instance-specific class labels \cite{Zhang2021}. 
These models can also learn meaningful cluster representations even in the overclustering scenario without knowledge of the underlying amount of clusters \cite{Hsu2016}.
The majority of research on constrained clustering focuses on the \textit{constrained} scenario, where each data point is associated with at least one constraint pair. 
As this setting still requires large sets of given constraints and hence incurs high annotation effort, we focus on the \textit{semi-constrained} setting where a clustering model is trained on both a small dataset of pairwise constraints and a large dataset of unconstrained samples.

\begin{figure}
    \centering
    \includegraphics[width=8.5cm]{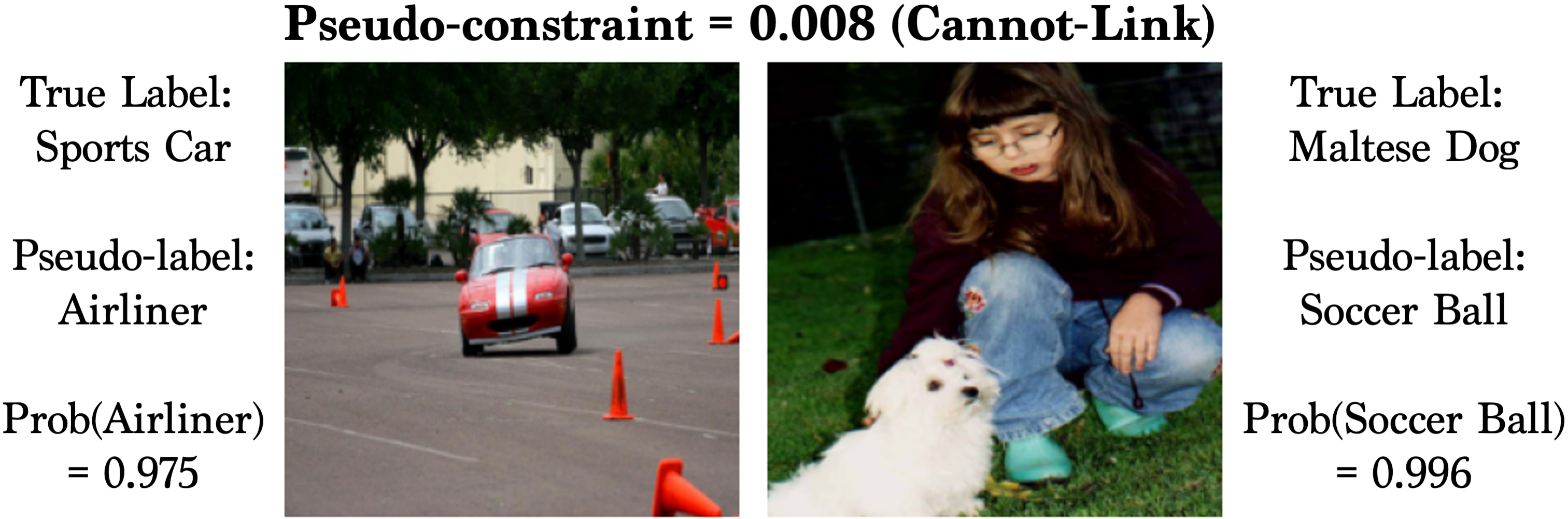}
    \caption{Illustration of pseudo-constraining. While the model creates overconfident, wrong pseudo-labels for both unlabeled samples, it still yields a semantically correct pseudo-constraint.}
    \label{fig:pc_motivation}
\end{figure}

While a great deal of progress has been made in semi-supervised learning when class labels are provided, we identify through analysis a number of challenges when applying such methods to the constrained clustering setting. 
One of the most effective methods in semi-supervised learning, pseudo-labeling, utilizes confident predictions on unlabeled data in training and is therefore prone to \textit{confirmation bias}. 
Specifically, unlabeled samples that were confidently assigned the wrong class label by the model are selected as pseudo-labels, which leads to subsequent model degradation \cite{Arazo2020}.
We analyze this issue in the context of constraint labels and propose a \textit{pseudo-constraining} mechanism that we show can mitigate it, by generating pseudo-constraints from the pseudo-labels (see Fig.~\ref{fig:pc_motivation}).
Further, we argue that a confidence-based pseudo-label selection criterion is inappropriate in this setting as it leads to the unnecessary de-selection of unconstrained samples that contain valuable information for subsequent pseudo-constraining.
We, therefore, propose an entropy-based criterion to select \textit{informative} unconstrained samples and show its superiority.
The combination of these two methods, \textit{ConstraintMatch}, facilitates effective pseudo-labeling and unifies the initial and auxiliary learning task. 
We show that ConstraintMatch is able to outperform several state-of-the-art baselines using only a few constraint annotations by substantial margins, even in the more challenging overclustering scenario.

\begin{figure*}
    \centering
    \includegraphics[width=11cm]{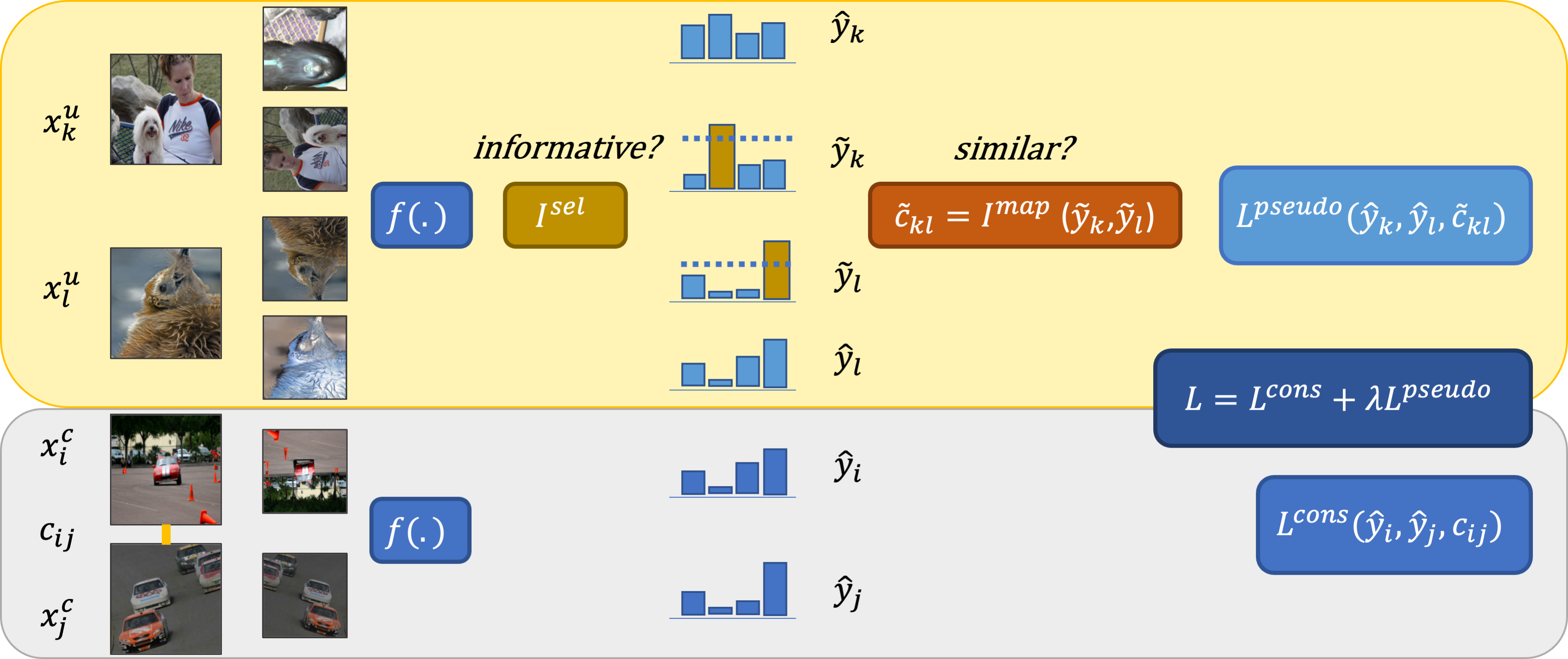}
    \caption{ConstraintMatch combines pairwise training on constrained (gray) and unconstrained (yellow) samples leveraging weak and strong data augmentations.
    The criterion $\mathcal{I}^{sel}$ is used to select \textit{informative} pseudo-labels from unconstrained samples which are then mapped to pairwise pseudo-constraints via $\mathcal{I}^{map}$ to overcome the \textit{confirmation bias}. 
    Predictions from model $f()$ over strongly augmented versions of these samples serve as inputs to the auxiliary loss $\mathcal{L}^{pseudo}$ to enforce consistency in predicted cluster assignments and the final model is trained on a combination of the pseudo-constrained and the constrained loss $\mathcal{L}^{cons}$.}
    \label{fig:ccm}
\end{figure*}

\textbf{Contributions} We 1) propose \textit{ConstraintMatch} as a method for semi-constrained training of clustering models leveraging a large set of unconstrained samples next to a small set of pairwise constraints. 
Within a series of experiments, we 2) specifically make the case for pseudo-constraints over naive pseudo-labels and provide a detailed analysis of ConstraintMatch's several components.
Furthermore, we 3) empirically prove the strong performance of ConstraintMatch of up to 16.75\% NMI over the constrained baseline on a series of five challenging benchmark datasets in both the regular and the overclustering scenario.
Thereby, we evaluate models in different settings to unify the evaluation of modern deep clustering approaches and 4) release our source code\footnote[1]{https://github.com/slds-lmu/constraintmatch} for future research on semi-constrained clustering.

%
%

\section{Related Work}
We provide an overview of the context of ConstraintMatch at the intersection of deep clustering, constrained clustering, and semi-supervised learning in the following.

\textbf{Deep Clustering} 
Early methods for deep clustering combine a reconstruction target with a clustering loss to learn expressive clustering features via reconstruction \cite{Guo2017, Li2018, Tian2017}. 
Subsequent approaches shift this focus toward low-level features via alternating cluster assignments with those provided by traditional clustering algorithms \cite{Yang2016, Caron2018}.
More recent research is directed at mapping the data onto low-dimensional representations which serve as a training target for similarity-based losses and as cluster predictions during model inference \cite{Chang2017, Wu2019, Ji2019, Li2021, Niu2021}.
Van Gansbeke et al. \cite{VanGansbeke2020} found these approaches are prone to learning low-level features which lack meaning for semantic clustering next to heavy dependence on network initialization.
Therefore, they propose SCAN as a two-step approach where feature representations learned via contrastive pretext tasks \cite{Chen2020, He2020} are used to mine nearest neighbors of the unlabeled samples. 
The model is then trained via a clustering loss which maximizes the alignment of their joint feature representations and enables clustering in the absence of the true underlying amount of clusters. 
SCAN was decidedly evaluated on test datasets only to prove its efficacy on new, unseen data. 
While this is appealing from a modeling perspective, it prevents direct comparison with prior work where clustering models are evaluated on the union of training and test datasets. 
We unify this model comparison in our experiments and find SCAN to perform on par with subsequent approaches TCC \cite{Shen2021}, CC \cite{Li2021}, and MICE \cite{Tsai2021}. 
Hence, we use SCAN as a starting point for ConstraintMatch. 

\textbf{Constrained Clustering} 
The introduction of binary instance-level constraints for clustering \cite{Wagstaff2000} led to the adaptation of existing clustering methods towards the use of constraints \cite{Wagstaff2001}, see \cite{Gancarski2020} for an overview. 
With the proposal of the KCL loss, Hsu et al. \cite{Hsu2016} introduced constrained clustering and overclustering to deep learning. 
They further showed its applicability to transfer learning \cite{Hsu2018a} and introduced the Meta-classification-likelihood (MCL) for improved model training with pairwise constraints \cite{Hsu2019} and both loss formulations can not be used with unconstrained data. 
Zhang et al. \cite{Zhang2021} provide a framework to work with various types of constraints.

\textbf{Semi-supervised Learning} In semi-supervised classification, the rationale of consistency regularization lead to substantial improvements over supervised baselines \cite{Sohn2020, Kuo2020, Berthelot2019, Zhang2021c}.
Among these, FixMatch \cite{Sohn2020} yields state-of-the-art model performance even in settings with very low supervision.
It combines confidence-based pseudo-labeling \cite{Lee2013} with consistency regularization using the weak-and strong augmentation scheme over unlabeled samples. 

\textbf{Semi-constrained Clustering} With S$^3$C$^2$, a two-stage approach was proposed that leverages pseudo-constraints mined from a siamese network trained on few constraints \cite{Smieja2020}. 
While this approach was shown to perform well on simple benchmarks, it lacks end-to-end training and requires the true amount of clusters as input.
Similarly, the approach by Fogel et al. \cite{Fogel2019} requires said amount of true clusters as input next to being a transductive method prohibiting inference on unseen data without access to the training data after training. 
PCOG \cite{Nie2021} is another transductive method that also requires the true amount of cluster while its spectral decomposition component hinders it from scaling to large datasets.
The approach of Shukla et al. \cite{Shukla2020} relies on a few class labels, rendering it non-applicable for scenarios where only constraints are present.
In contrast to these approaches, we introduce a semi-constrained clustering method that only relies on constraint annotations, leverages unconstrained data, is inductive, and works well without knowledge w.r.t the true amount of clusters.

%
%

\section{Method}
\label{ch3:motivation}

\subsection{Notation}

We consider a dataset $\mathcal{D}$ which consists of  constrained and unconstrained datasets $\mathcal{D}^c$ and $\mathcal{D}^u$. 
$\mathcal{D}^c$ contains $n_c$ constrained pairs of the form $x_{ij}^c=(x_i^{c}, x_j^{c}, c_{ij}) \in \mathcal{D}^c$ where $x_i^{c}, x_j^{c}$ refer to two input samples and $c_{ij}\in \{0, 1\}$ to the associated binary constraint.
These constraints describe that both samples either correspond to the same cluster $c_{ij}=1$, \textit{Must-Link} constraints (ML), or to different clusters $c_{ij}=0$, \textit{Cannot-Link} constraints (CL).
$\mathcal{D}^u$ consists of $n_u$ unconstrained input samples $x_i^{u} \in \mathcal{D}^u$.
We denote $\mathcal{B} \subset \mathcal{D}, \mathcal{B}^c \subset \mathcal{D}^c, \mathcal{B}^u \subset \mathcal{D}^u$ as batches of input samples $x_i$ of the respective datasets. 
We refer to true class labels as $y_i \in \mathcal{Y}$ where $K=|\mathcal{Y}|$ describes the amount of true classes, i.e. the amount of underlying clusters $K$, in the dataset.
Note that when $K$ is not known, the model may have a different number of outputs $n_{out}$ than the ground truth number of clusters. 
We aim at training a clustering model $f$ in the form of a neural network with its final head consisting of $n_{out}$ output neurons followed by a softmax layer, i.e. the model predicts a probability distribution over cluster assignments $\hat y_i = f(x_i)$ where $\hat y_{il}$ denotes the predicted probability of $x_i$ belonging to cluster $l \in 1, ...,n_{out}$.
Similarly, we refer to pseudo-labels as $\tilde y_i$ and to pseudo-constraints as $\tilde c_{ij} \in [0, 1]$. 
We introduce the criterion $\mathcal{I}^{sel}$ which selects a subset of informative pseudo-labels $\tilde y_i$ based on their predicted cluster assignments $\hat y_i$ from $\mathcal{B}$. 
From pairs of selected pseudo-labels $\tilde y_i, \tilde y_j$ we construct pseudo-constraints $\tilde c_{ij}$ using a second criterion $\mathcal{I}^{map}$. 

\subsection{Algorithm}
\label{ch3:algo}

\textit{ConstraintMatch} is an annotation-efficient method that can leverage large unconstrained (i.e. unlabeled) data $\mathcal{D}^u$ next to few constraint pairs $\mathcal{D}^c$ to train a clustering model $f$.
It uses unsupervised clustering in a pretraining step and combines training strategies from constrained clustering \cite{Hsu2016, Hsu2019} with the semi-supervised method Fixmatch \cite{Sohn2020}, refer to Fig.~\ref{fig:ccm} for illustration. 
We use SCAN \cite{VanGansbeke2020} for the pretraining step but other pretraining methods would also be applicable. 
Specifically, pseudo-labeling (i.e. self-training) has proven itself an effective method for leveraging unlabeled data and is a key component of recent semi-supervised classification models \cite{Sohn2020, Kuo2020}. 
In naive pseudo-labeling, confident model predictions over unlabeled samples are used as pseudo-targets in an auxiliary classification loss to guide model training next to the initial supervised loss, assuming that model confidence is associated with model correctness \cite{Lee2013, VanEngelen2020}.
Adapting this concept to constrained clustering, we identified three main weaknesses which we overcome: 1) \textbf{Pseudo-constraining:} Prediction errors in the selected pseudo-labels can amplify during training, potentially leading to model degradation, also known as \textit{confirmation bias}~\cite{Arazo2020}. 
We, therefore, propose the generation of \textit{pseudo-constraints}, as pairwise constraints result in a simpler problem reduction~\cite{Hsu2019}. 
2) \textbf{Informativeness criterion} to carry information of whether two samples $x_i$ and $x_j$ are predicted to be in the same or a different cluster, which cannot be done via maximal prediction probability \cite{Sohn2020} or alternative uncertainty metrics \cite{Arazo2020}, and 3) \textbf{Unification of losses} utilizing a constraint-based loss for the unlabeled set.  


Our overall algorithm processes unconstrained batches $\mathcal{B}^u$ via an unconstrained branch and constrained batches $\mathcal{B}^c$ within a constrained branch to enable training of clustering model $f$ in this semi-constrained data scenario. 
The constrained branch is trained via a pairwise objective $\mathcal{L}^{cons}$ which allows the training of the model $f$ on binary pairwise constraints $x_{ij}^c=(x_i^c, x_j^c, c_{ij}) \in \mathcal{D}^c$.
Therefore, we combine the predictions from model $f$  over weakly augmented constrained samples $a(x_i^{c}), a(x_j^{c})$ along the associated constraint $c_{ij}$ within the pairwise loss function $\mathcal{L}^{cons}$. 
For the unconstrained branch, we build upon the intuition of consistency regularization via weak and strong data augmentation strategies $a()$ and $A()$ \cite{Sohn2020} as follows. 
Given a pair of unconstrained samples $x_i^{u}, x_j^{u}$, we use the selection criterion $\mathcal{I}^{sel}$ to select \textit{informative} model predictions over weakly augmented versions of those samples $(f(a(x_i^{u})), f(a(x_j^{u})))$ as pseudo-labels $(\tilde y_i, \tilde y_j)$.
These are then combined into pseudo-constraints $\tilde c_{ij}$ via $\mathcal{I}^{map}$ and used as targets within the auxiliary loss function $\mathcal{L}^{pseudo}$. 
Model predictions over strongly augmented versions of the unconstrained pair $(\hat y_i, \hat y_j)=(f(A(x_i^u)), f(A(x_j^u)))$ serve as inputs for $\mathcal{L}^{pseudo}$. 
ConstraintMatch is trained using the combined loss function $\mathcal{L}=\mathcal{L}^{cons}+\lambda \mathcal{L}^{pseudo}$. 
The components of ConstraintMatch are explained in the following.

\textbf{1) Pseudo-Label Selection} \label{ch3:PL}
Semi-supervised approaches use model confidence as measured via the maximal prediction probability \cite{Sohn2020} or alternative uncertainty metrics \cite{Arazo2020} as selection criteria.
Model confidence assumes uni-modal model predictions, i.e. the model is confident that sample $x_i$ belongs to class $\hat y_i=l$. 
In contrast to pseudo-labeling, we do not need the information of whether one sample $x_i$ is confidently predicted to be in class $\hat y_i=l$ but the information of whether samples $x_i$ and $x_j$ are predicted to be in the same or a different cluster.
Filtering for model confidence de-selects multi-modal model predictions that would, for instance, be assigned to two clusters with high probability each - pseudo-constraining allows us to use such multi-modal predictions.

Given a batch of model predictions over weakly-augmented, unconstrained samples, we aim to select those that are important for the subsequent pseudo-constraint generation.
We propose measuring such \textit{informativeness} of a probability vector $\hat y_i$ using the \textit{normalized entropy}:

\begin{equation}
    \mathcal{H}_n(\hat y_i)=-\frac{1}{\textrm{log}(n_{\textrm{out}})}\sum_{l=1}^{n_{out}}p(\hat y_{il})\textrm{log}(p(\hat y_{il}))
\end{equation}

\noindent with $\mathcal{H}_n(\hat y_i) \in [0; 1]$ where $\mathcal{H}_n(\hat y_i)=1$ describes the minimum level of information and maximal entropy and $\mathcal{H}_n(\hat y_i)=0$ the maximum level of potential informativeness and minimal entropy where the model places the entire probability mass in one cluster. 
Hence, we use the normalized entropy in combination with a threshold hyperparameter $\tau \in [0; 1]$ as criterion to select pseudo-labels:

\begin{equation}
    \mathcal{I}_{\tau}^{sel}(\hat y_i)=\mathbbm{1}(\mathcal{H}_n(\hat y_i)<\tau)
\end{equation}

\noindent where $\mathbbm{1}$ is an indicator function. 
In the experiment section, we provide an empirical analysis of the suitability of this criterion next to a sensitivity analysis of $\tau$.

\looseness=-1 \textbf{2) Pseudo-Constraining} \label{ch3:PC}
Confirmation bias is a critical problem in pseudo-labeling methods \cite{Arazo2020}, and in constraint-based clustering, there is an opportunity to alleviate this. 
As an illustration, refer to the two unconstrained samples $x_i^u, x_j^u$ from Fig.~\ref{fig:pc_motivation}: the true label $y_i$ of $x_i^u$ would be "sports car" while the model wrongly but confidently assigns it to the "airliner" cluster and similarly $x_j^u$ is assigned the wrong cluster "soccer ball" instead of the true $y_j$ "maltese dog" (see Fig.~\ref{fig:failed_pls} in the Appendix for more examples). 
While this prediction error would lead to a wrong prediction target in naive pseudo-labeling and hence confuse model training, the resulting pseudo-constraint $\tilde c_{ij}=0.008$ would still be correctly assigned as Cannot-Link, as $\tilde y_j \neq \tilde y_i$ in both situations. 
Therefore, we create pseudo-constraints from the pseudo-labels to drive the loss function $\mathcal{L}^{pseudo}$. 
Given a batch of informative pseudo-labels, we next combine pseudo-label pairs $\tilde y_i, \tilde y_j$ into pseudo-constraints $\tilde c_{ij}$, expressing the (dis)-similarity of those samples. 
As $\tilde y_i, \tilde y_j$ are probability vectors, we propose to use a divergence measure to quantify this distance and derive a meaningful pseudo-constraint. 
The Jensen-Shannon-Distance \cite{Lin1991} allows the symmetric mapping of two probability vectors onto a similarity score: 

\begin{equation}
    JSD(\tilde y_i, \tilde y_j) = \sqrt{((KL(\tilde y_i|m) + KL(\tilde y_j|m))/2}
\end{equation}

\noindent where $m=(\tilde y_i + \tilde y_j) / 2$ and $KL(y_i|m)$ refers to the Kullback-Leibler Distance between $\tilde y_i$ and $m$ and $JSD(\tilde y_i, \tilde y_j) \in [0, 1]$. 
We exploit this property and use the inverse Jensen-Shannon-Distance to calculate soft pseudo-constraints $\tilde c_{ij} = 1-JSD(\tilde y_i, \tilde y_j) \in [0;1]$ where $\tilde c_{ij}=0.0$ resembles a Cannot-Link and $\tilde c_{ij}=1.0$ a Must-Link pseudo-constraint over all pairwise combinations of the informative pseudo-labels.
We refer to this inverse Jensen-Shannon-Distance as $\mathcal{I}^{map}(\tilde y_i, \tilde y_j)$ in Fig.~\ref{fig:ccm}.
Pseudo-constraints are generated over the combined batch $\mathcal{B} = \mathcal{B}^c\cup\mathcal{B}^u$, treating the samples in $\mathcal{B}^c$ as unconstrained. 

\looseness=-1 \textbf{3) Pairwise Loss Function }
\label{ch3:Loss} There exists a variety of loss functions that can deal with pairwise constraints \cite{Zhang2021} with the KCL \cite{Hsu2016} and the MCL \cite{Hsu2019} being the most prominent ones. 
Following the findings of Hsu et al. \cite{Hsu2019} and guided by preliminary experimental results, we propose the use of the MCL as a pairwise loss function, as it was shown to result in higher model performance and smoother model training, and is hyperparameter-free.
The MCL is aligned on the binary cross-entropy loss and follows the definition:

\begin{equation}
    \mathcal{L}(c_{ij}, \hat c_{ij}) = - \sum_{ij}c_{ij} \textrm{log}(\hat c_{ij}) + (1-c_{ij}) \textrm{log}(1-\hat c_{ij})
\end{equation}

\noindent where $\hat c_{ij} = \left<\hat y_i, \hat y_j\right>$ combines the individual predicted cluster assignment vectors into an alignment score and $c_{ij} \in \{0, 1\}$ refers to the pairwise constraint $c_{ij}=0$ for Cannot-Link and $c_{ij}=1$ for Must-Link constraints. 
The MCL allows training with soft constraints $c_{ij}\in [0,1]$, similar to the use of soft labels in the cross-entropy loss \cite{Meister2020}. 
We use this property for the processing of soft pseudo-constraints $\tilde c_{ij}$ as explained above and analyzed further in the experiments section.
ConstraintMatch is trained on the combined loss: 

\begin{equation}
\begin{split}
    \mathcal{L} = &\sum_{x_i, x_j \in \mathcal{B}^c} \mathcal{L}^{cons}\left( c_{ij}, \left<f(a(x_i)),f(a(x_j))\right> \right) +\\
    &\lambda\sum_{x_k, x_l \in \mathcal{B}} \mathcal{L}^{pseudo}\left( \tilde c_{kl}, \left<f(A(x_k)),f(A(x_l))\right> \right)
\end{split}
\end{equation}

\noindent where hyperparameter $\lambda \geq 0$ controls the impact of the pseudo-constraint loss and is tuned on the validation set.

%
%

\section{Experiments}
\label{ch4}
In this section, we compare the performance of ConstraintMatch with prior work on five challenging benchmark datasets and provide empirical evidence for the effectiveness of pseudo-constraining.
This includes i) the relative improvement of ConstraintMatch across various baselines, ii) an empirical analysis of the benefit of pseudo-constraining and iii) its robustness w.r.t annotation noise, iv) analyses of the algorithmic choices made for its several components, and v) an evaluation of ConstraintMatch in the overclustering scenario. 

\begingroup
\setlength{\tabcolsep}{2pt}

\begin{table*}[ht]
\caption{Comparison of ConstraintMatch with relevant baselines (B), competitors (C), and upper bound models (U) across datasets and varying amounts of constraints $n_c$.
Performance metrics were averaged over five folds and calculated on separate test splits in the upper part and in the Train(+Test) setting in the lower part. 
Best results comparing ConstraintMatch with the baseline and competitor models are shown in bold and $^\dagger$ denotes values reported in the literature. 
Statistical significance for differences in model performance between ConstraintMatch and the constrained competitor for $n_c \in \{\text{5k}, \text{10k}\}$ respectively established using the Wilcoxon signed-rank test \cite{Demsar2006, Wilcoxon1945} (significance code $^{*}:p < 0.05$).
}
\label{tab:main}
\centering
\resizebox{18cm}{!}{
\begin{tabular}{@{} l l c l@{\hskip 0.25cm} l l l@{\hskip 0.25cm} l l l@{\hskip 0.25cm} l l l@{\hskip 0.25cm} l l l@{\hskip 0.25cm} l l l}
\toprule
& & & & \multicolumn{3}{c}{Cifar 10} & \multicolumn{3}{c}{Cifar 100-20} & \multicolumn{3}{c}{STL 10} & \multicolumn{3}{c}{ImageNet 10} & \multicolumn{3}{c}{ImageNet Dogs} \\
\cmidrule(l{2pt}r{2pt}){5-7} \cmidrule(l{2pt}r{2pt}){8-10} \cmidrule(l{2pt}r{2pt}){11-13} \cmidrule(l{2pt}r{2pt}){14-16} \cmidrule(l{2pt}r{2pt}){17-19}
Split & Model & & $n_c$ & ACC & NMI & ARI & ACC & NMI & ARI & ACC & NMI & ARI & ACC & NMI & ARI & ACC & NMI & ARI\\ 
\midrule
Test & Supervised$^\dagger$ & U & & 93.80 & 86.20 & 87.00 & 80.00 & 68.00 & 63.20 & 80.60 & 65.90 & 63.10 & - & - & - & - & - & -\\
& Fully Constrained & U & & 94.86 & 88.39 & 89.11 & 77.99 & 68.37 & 61.94 & 90.49 & 80.94 & 80.47 & 96.64  & 93.45 & 92.60 & 67.10 & 73.52 & 58.13\\
\cmidrule(l{2pt}r{2pt}){2-19}
& SCAN$^\dagger$ \cite{VanGansbeke2020} & B & 0 & 87.60 & 78.70 & 75.80 & 45.90 & 46.80 & 30.10 & 76.70 & 68.00 & 61.60 & 86.20 & 81.57 & 75.71 & 47.20 & \textbf{55.42} & 35.87 \\
\\
& Constrained & C & 5k & 90.12 & 80.52 & 80.02 & 50.99 & 46.23 & 32.63 & 85.90 & 74.62 & 72.51 & 93.12 & 87.43 & 85.49 & 44.08 & 43.27 & 28.92 \\
& ConstraintMatch & & 5k & 92.23$^{*}$ & 84.64$^{*}$ & 84.27$^{*}$ & 54.19$^{*}$ & 52.74$^{*}$ & 37.84$^{*}$ & 88.20$^{*}$ & 78.20$^{*}$ & 76.68$^{*}$ &  94.68$^{*}$ & 90.43$^{*}$ & 88.61$^{*}$ & 49.43$^{*}$ & 55.23$^{*}$ & 38.12$^{*}$ \\
& Constrained & C & 10k & 90.89 & 81.73 & 81.47 & 52.45 & 46.57 & 33.79 & 88.21 & 77.63 & 76.38 & 94.90 & 90.42 & 89.04 & 45.52 & 44.44 & 30.10 \\
& ConstraintMatch & & 10k & \textbf{93.17$^{*}$} & \textbf{85.88}$^{*}$ & \textbf{85.92}$^{*}$ & \textbf{57.15}$^{*}$ & \textbf{54.37}$^{*}$ & \textbf{40.59}$^{*}$ & \textbf{90.08}$^{*}$ & \textbf{80.57}$^{*}$ & \textbf{79.81}$^{*}$ & \textbf{95.68} & \textbf{92.09} & \textbf{90.70} & \textbf{50.73}$^{*}$ & 54.92$^{*}$ & \textbf{38.34}$^{*}$ \\
\midrule
Train & PICA$^\dagger$ \cite{Huang2020}       & B & 0 & 69.60 & 59.10 & 51.20 & 33.70 & 31.00 & 17.10 & 71.30 & 61.10 & 53.10 & 87.00 & 80.20 & 76.10 & 35.20 & 35.20 & 20.10 \\
(+Test) & MICE$^\dagger$ \cite{Tsai2021}        & B & 0 & 83.50 & 73.70 & 69.80 & 44.00 & 43.60 & 28.00 & 75.20 & 63.50 & 57.50 & - & - & - & 43.90 & 42.30 & 28.60 \\
& CC$^\dagger$ \cite{Li2021}            & B & 0 & 79.00 & 70.50 & 63.70 & 42.90 & 43.10 & 26.60 & 85.00 & 76.40 & 72.60 & 89.30 & 85.90 & 82.20 & 42.90 & 44.50 & 27.40 \\
& TCC$^\dagger$ \cite{Shen2021}         & B & 0 & 90.60 & 79.00 & 73.30 & 49.10 & 47.90 & 31.20 & 81.40 & 73.20 & 68.90 & 89.70 & 84.80 & 82.50 & \textbf{59.50} & \textbf{55.40} & \textbf{41.70} \\
& SCAN \cite{VanGansbeke2020} & B & 0 & 88.53 & 80.09 & 77.72 & 50.67 & 47.72 & 33.07 & 81.28 & 70.15 & 65.22 & 91.63 & 84.00 & 82.93 & 44.06 & 45.09 & 30.75 \\
\\
& Constrained                 & C & 5k & 91.14 & 82.30 & 82.05 & 51.63 & 46.58 & 33.37 & 80.51 & 68.38 & 63.68 & 95.09 & 88.42 & 89.49 & 43.25 & 38.82 & 28.93 \\
& ConstraintMatch             & & 5k & 92.67$^{*}$ & 85.12$^{*}$ & 84.98$^{*}$ & 54.16$^{*}$ & 52.68$^{*}$ & 37.79$^{*}$ & 82.97$^{*}$ & 71.13$^{*}$ & 67.80$^{*}$ & 95.61$^{*}$ & 89.64$^{*}$ & 90.59$^{*}$ & 47.63$^{*}$ & 47.82$^{*}$ & 35.95$^{*}$ \\
& Constrained                 & C & 10k & 92.21 & 83.07 & 84.08 & 53.13 & 47.20 & 34.86 & 89.90 & 80.12 & 79.49 & 96.47 & 91.18 & 92.36 & 44.17 & 40.07 & 30.13 \\
& ConstraintMatch             & & 10k & \textbf{93.61}$^{*}$ & \textbf{86.55}$^{*}$ & \textbf{86.80}$^{*}$ & \textbf{57.18}$^{*}$ & \textbf{53.37}$^{*}$ & \textbf{40.30}$^{*}$ & \textbf{91.30}$^{*}$ & \textbf{82.42}$^{*}$ & \textbf{82.13}$^{*}$ & \textbf{96.68}$^{*}$ & \textbf{91.59}$^{*}$ & \textbf{92.80}$^{*}$ & 49.34$^{*}$ & 49.16$^{*}$ & 37.37$^{*}$ \\
\bottomrule
\end{tabular}
}
\end{table*}
\endgroup

\subsection{Experimental Setup}
\label{ch4:experimentalsetup}

\textbf{Datasets and Constraint Mining} We use the Cifar10 \cite{Krizhevsky2009}, Cifar100 \cite{Krizhevsky2009}, STL10 \cite{Coates2011}, ImageNet-10 \cite{Chang2017} and ImageNet-Dogs \cite{Chang2017} datasets to demonstrate the effectiveness of the proposed method.
We use the 20 superclasses in Cifar100 as ground truth labels for constraint mining following prior work \cite{VanGansbeke2020, Li2021}, declaring it as Cifar100-20 in the following.
We adhere to the provided train/test splits to enable comparison with prior work evaluated on separate test datasets \cite{VanGansbeke2020, Hsu2019}. 
Recent deep clustering approaches instead are evaluated on the training set or the union of training and test datasets \cite{Shen2021, Tsai2021, Li2021}.  
To be comparable with both bodies of literature, we provide benchmark results in both settings marked as "Test" and "Train(+Test)" in Table~\ref{tab:main} following \cite{Li2021}.
An overview of the used datasets, the amount of sampled constraints as well as the training and validation splits for hyperparameter tuning is provided in Table~\ref{tab:datalarge} in the Appendix. 
For constraint-sampling, we randomly sample $n_c$ constraints from each dataset via the following procedure: $n_c$ samples are randomly sampled without replacement as constraint members $(x_i^c, y_i)$ and for each of those samples, a second pair member $(x_j^c, y_j)$ is randomly chosen with replacement from the remaining training samples to create the constraint pair $(x_i^c, x_j^c, c_{ij})$, where $c_{ij}=1$, if $y_i=y_j$ and $c_{ij}=0$, if $y_i \neq y_j$. 
This results in a dataset $\mathcal{D}^c$ of $n_c$ constrained samples $(x_i^c, x_j^c, c_{ij})$.
To account for randomness in the constraint sampling process, we report performance averaged over five random sampling repetitions. 

\textbf{Implementation Details} In accordance with prior work \cite{VanGansbeke2020}, we used a ResNet-18 backbone architecture \cite{He2016} for the experiments with the Cifar10, Cifar100-20, and STL10 datasets and a ResNet-34 backbone \cite{He2016} following \cite{Li2021} for the ImageNet datasets.
We used model weights that were pre-trained via SCAN \cite{VanGansbeke2020} for the initialization of the model backbone as ConstraintMatch benefits from expressive feature representations as a warm start.
Next to the model weights released by \cite{VanGansbeke2020} for Cifar10, Cifar100-20, and STL10 we used the authors' codebase\footnote[2]{https://github.com/wvangansbeke/Unsupervised-Classification} to pretrain the ResNet-34 backbone via SCAN and then used these resulting model weights for model initialization.
For model training, we used a standard SGD optimizer with momentum set to $0.9$ and weight decay regularization \cite{Sutskever2013} and all models were trained for a total of $20000$ optimization steps unless noted otherwise. 
We used a cosine learning rate scheduler \cite{Loshchilov2017} which updates the learning rate at each update step to $\eta \textrm{ cos}\left(\frac{7 \pi t }{16T}\right)$ with $\eta$ being the initial learning rate, $t$ the current training step and $T=20000$ the total amount of training steps following \cite{Sohn2020}.
Hyperparameters were tuned via a grid search on constraints mined from the validation datasets with hyperparameter ranges shown in Table~\ref{tab:hpars} and more details on the validation splits are given in Table~\ref{tab:datalarge} in the Appendix.
The size of constrained/unconstrained batches was set to $200/ 600$ respectively for Cifar10 and Cifar100-20 and to $100/ 300$ for the other datasets. 

\textbf{Model Comparison} We compare ConstraintMatch with different baselines (B), competitors (C), and upper bound models (U). 
This includes deep clustering models SCAN~\cite{VanGansbeke2020}, TCC~\cite{Shen2021}, CC~\cite{Li2021}, MICE~\cite{Tsai2021} and PICA~\cite{Huang2020} as baselines and a constrained clustering model that was trained on $D^c$ using the MCL~\cite{Hsu2019} as competitor. 
As upper bounds, we compare with a fully constrained clustering model trained on a fully constraint version of training dataset $D$ and a supervised baseline where the backbone was trained on the fully labeled training set $D$ as reported by \cite{VanGansbeke2020}. 
The authors of MICE \cite{Huang2020}, PICA \cite{Tsai2021}, CC \cite{Li2021} and TCC \cite{Shen2021} used a ResNet-34 backbone for the Cifar10, Cifar100-20 and STL10 datasets. 
We used a ResNet-18 backbone for these three datasets in adherence with SCAN \cite{VanGansbeke2020}.
A comparison with the semi-constrained approaches was not possible due to a lack of open-source code and performance metrics on established benchmarks. 

\subsection{Results}
\label{ch4:results}

We summarize our main results in Table~\ref{tab:main} measuring model performance in Accuracy (ACC), Normalized Mutual Information (NMI) \cite{Strehl2002} and the Adjusted Rand Index (ARI) \cite{Steinley2004}, as standard in (constrained) clustering \cite{Zhang2021}.
As established in the literature \cite{Shen2021, VanGansbeke2020, Hsu2016}, we use the Hungarian Assignment method to optimally map the resulting cluster predictions to the true cluster labels \cite{Kuhn1955}.
As expected and shown in previous work \cite{Hsu2016, Hsu2019}, we find training with pairwise constraints to be a valid option to train strong-performing clustering models. 
This benchmark is the first attempt to compare SCAN with subsequent deep clustering methods on the union of train and test datasets showing that SCAN is competitive with those methods (lower part of Table~\ref{tab:main}). 
Further, fine-tuning SCAN via a subset of constraints improves model performance across all datasets but the fine-grained ImageNet-Dogs dataset.
Overall, ConstraintMatch outperforms the (un-)constrained baselines and competitors across all datasets in both evaluation settings except ImageNet-Dogs, a task with semantically very similar classes, where it falls behind TCC in the Train(+Test) evaluation. 
This fine-grainedness makes training of constrained clustering models challenging, as the distinction between Must- and Cannot-link loses expressiveness which also explains the worse performance of the constrained competitor compared to SCAN.
We interpret the finding that ConstraintMatch, in turn, outperforms both models by substantial margins in ACC and ARI as further proof of the effectiveness of pseudo-constraining. 
Relative (absolute) performance gains are the largest for Cifar100-20 with ConstraintMatch increasing model performance over the constrained baseline with 10k constraints by $8.96\%$ ($4.70$ percentage points) Accuracy, $16.75\%$ ($7.80$pp) NMI and $20.12\%$ ($6.80$pp) ARI on the test dataset. 
We attribute these large performance gains to the complexity of the task and the efficient use of pseudo-constraints in this complex 20-cluster setting.
Further, both the constrained competitor and ConstraintMatch benefit from more constraints $n_c$, with a larger relative performance increase for ConstraintMatch.
We yield a statistically significant difference in model performance for ConstraintMatch and the constrained competitor across all datasets for all performance metrics using a Wilcoxon signed-rank test \cite{Demsar2006, Wilcoxon1945} ($p < 0.05$) for $n_c \in \{\text{5k}, \text{10k}\}$.
Those empirical results confirm ConstraintMatch as a suitable method for semi-constrained clustering.

\subsection{The Empirical Case for Pseudo-Constraints}
\label{ch4:pseudoconstraints}

\begin{figure}
    \centering
    \includegraphics[width=7cm]{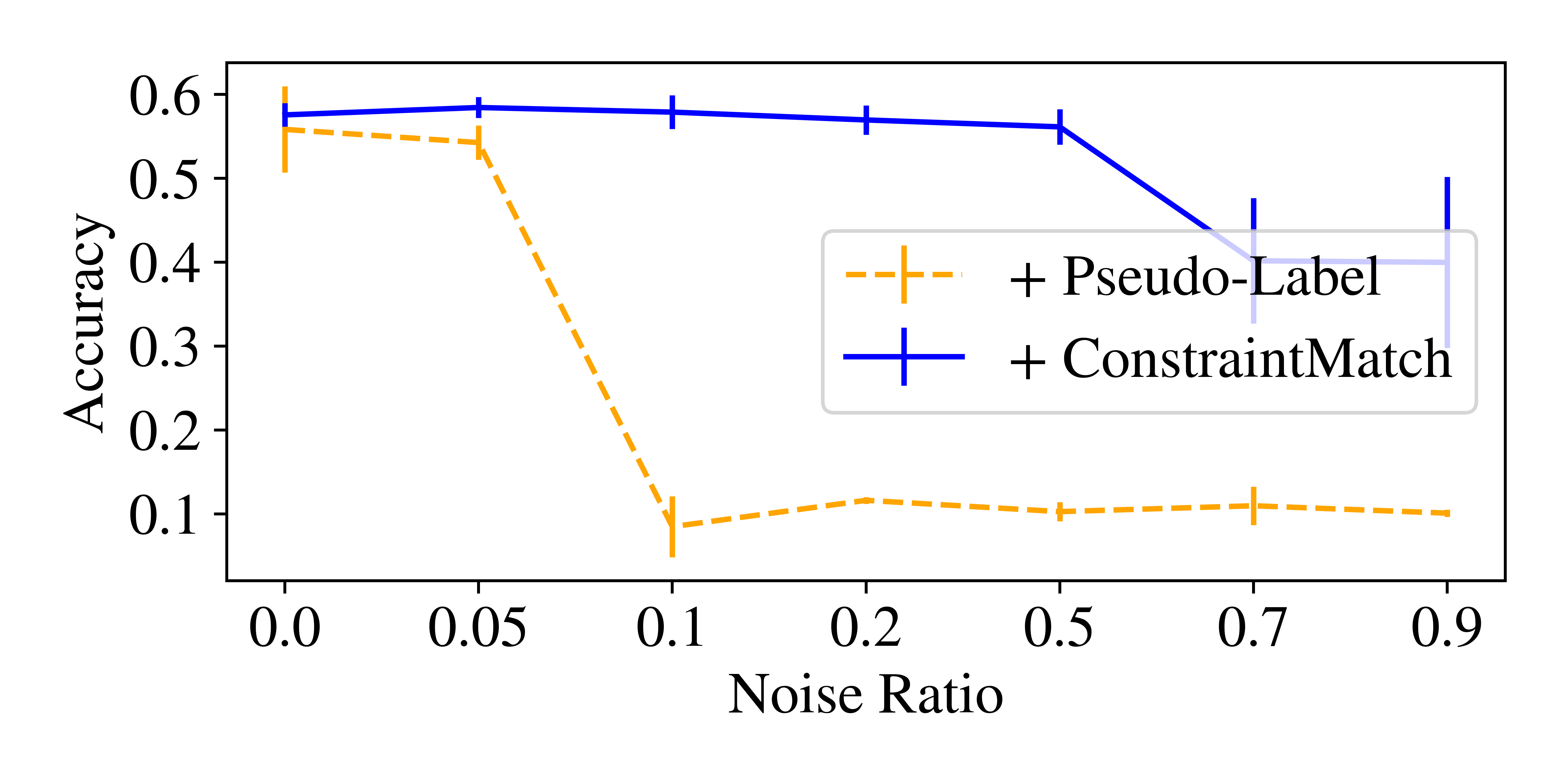}
    \caption{Robustness of the pseudo-labeling baseline and ConstraintMatch towards pseudo-label noise.}
    \label{fig:pcpl}
\end{figure}

We propose pseudo-constraining to overcome \textit{confirmation bias}. 
To support this claim, we conducted a simulation experiment to evaluate the robustness of naive pseudo-labeling against confirmation bias in comparison to subsequent pseudo-constraining within ConstraintMatch.
This naive pseudo-labeling baseline differs from ConstraintMatch in the handling of unconstrained samples, similar to the processing of unlabeled data in FixMatch \cite{Sohn2020}: weakly augmented, unconstrained samples are selected via a confidence threshold over their predicted cluster assignment and the major predicted cluster is chosen as pseudo-label.
Predictions over strong augmented versions of these samples then serve as input for an auxiliary cross-entropy loss function (see Fig.~\ref{fig:baseline_npl_cm} in the Appendix).
We introduce a mode-flip function $m(\hat y_i)$ that swaps the position of the two largest predicted probabilities within the model prediction $\hat y_i$. 
This simulates a prediction error where the model "confuses" two cluster assignments within the pseudo-labeling of $x_i$. 
We randomly apply $m()$ to a noise fraction $\rho$ of the unconstrained samples $x_i \in \mathcal{B}^u$ and train both models in this setting.
The results in Fig.~\ref{fig:pcpl} confirm our intuition as naive pseudo-labeling already degrades at $\rho\geq0.1$ while ConstraintMatch can cope with $\rho\leq0.5$. 

\begin{table}
    \caption{Ablation study on ConstraintMatch, results averaged over 5 folds with $n_c=10000$.}
    \label{tab:label_v_constraints}
    \centering
    \begin{tabular}{l c c c}
        \toprule
        Model & \multicolumn{3}{c}{Test Performance} \\
        & ACC & NMI & ARI \\
        \midrule
        SCAN & 45.90 & 46.80 & 30.10 \\
        + Constrained & 52.45 & 46.57 & 33.79 \\ 
        + Pseudo-Labeling & 55.38 & 53.49 & 39.98 \\ 
        \textbf{+ Pseudo-Constraining} & \textbf{57.15} & \textbf{54.37} & \textbf{40.59}\\
        \bottomrule
    \end{tabular}
\end{table}

Pseudo-constraining further allows to use the same pairwise loss function as both the auxiliary and the initial objective for model training.
We provide an ablation study on Cifar100-20 to quantify this benefit where we subsequently add constrained training, naive pseudo-labeling, and finally pseudo-constraining to the SCAN model, each with tuned hyperparameters.
The results in Table~\ref{tab:label_v_constraints} show that while the use of naive pseudo-labeling leads to a substantial performance gain over the constrained baseline, the subsequent application of pseudo-constraining within ConstraintMatch enables further model improvements. 
We conclude its effectiveness is grounded in both the robustness w.r.t. the confirmation bias and the similarity in training objectives.

\subsection{Additional Analyses}
\label{ch4:analyses}

\begin{figure*}[htbp]
    \centering
    \begin{subfigure}[b]{0.22\textwidth}
        \centering
        \captionsetup{width=.8\linewidth, size=scriptsize}
        \includegraphics[width=\textwidth]{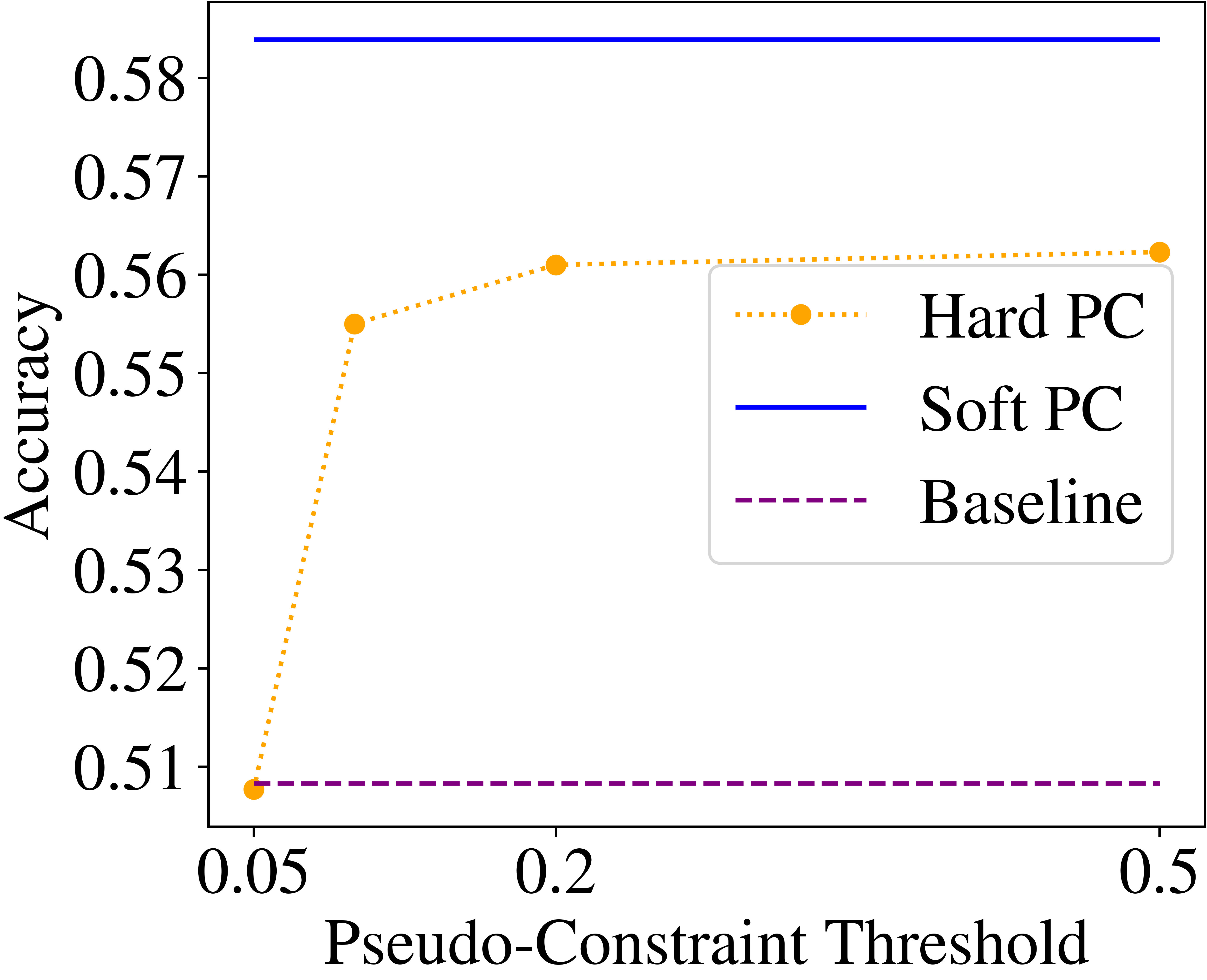}
        \caption{$\mathcal{I}^{map}$: soft vs. hard pseudo-constraints.}
        \label{fig:analysis1}
    \end{subfigure}
    \begin{subfigure}[b]{0.22\textwidth}
        \captionsetup{width=.8\linewidth, size=scriptsize}
        \includegraphics[width=\textwidth]{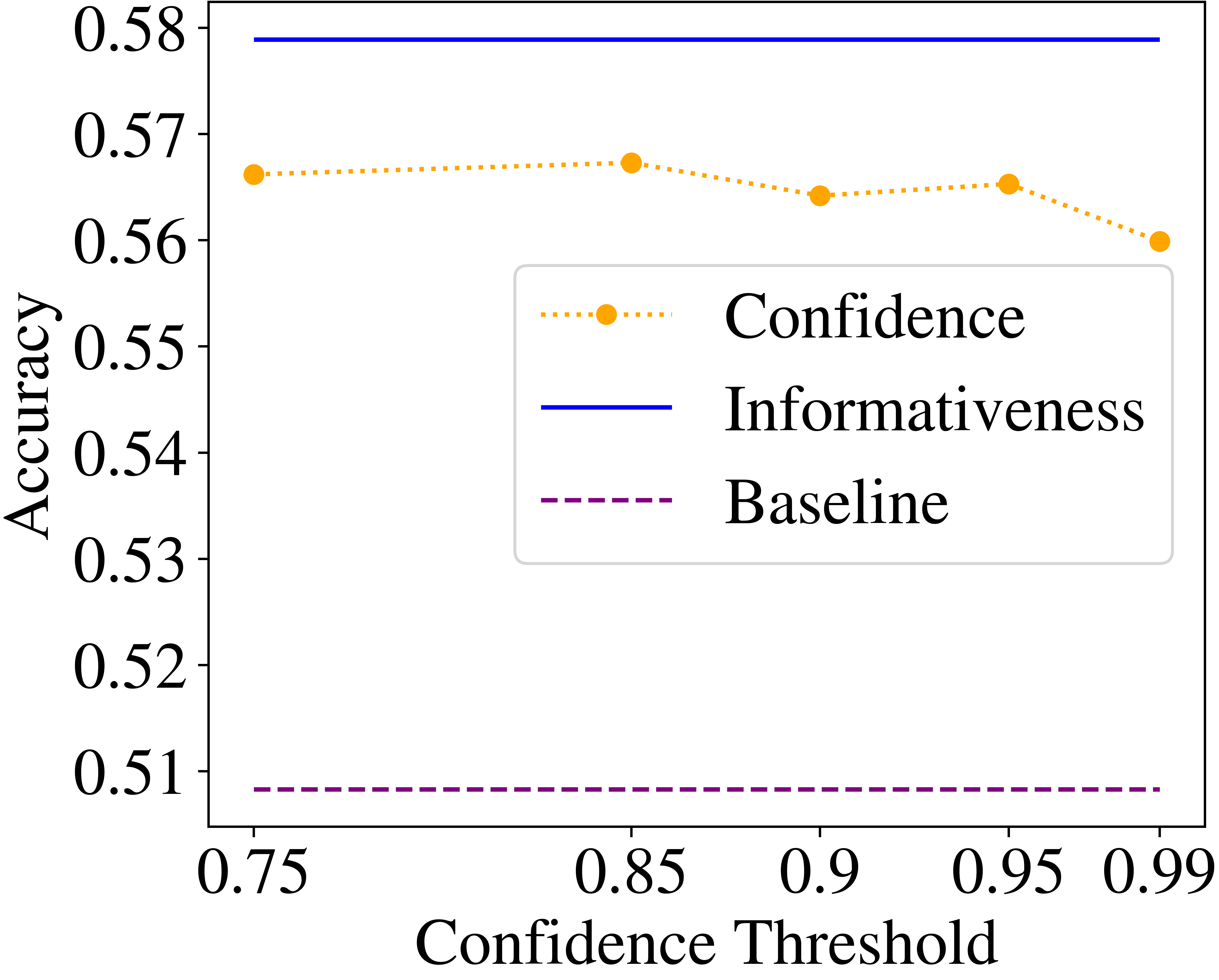}
        \caption{$\mathcal{I}^{sel}$: informativeness vs. confidence.}
        \label{fig:analysis2}
    \end{subfigure}
    \begin{subfigure}[b]{0.22\textwidth}
        \captionsetup{width=.7\linewidth, size=scriptsize}
        \includegraphics[width=\textwidth]{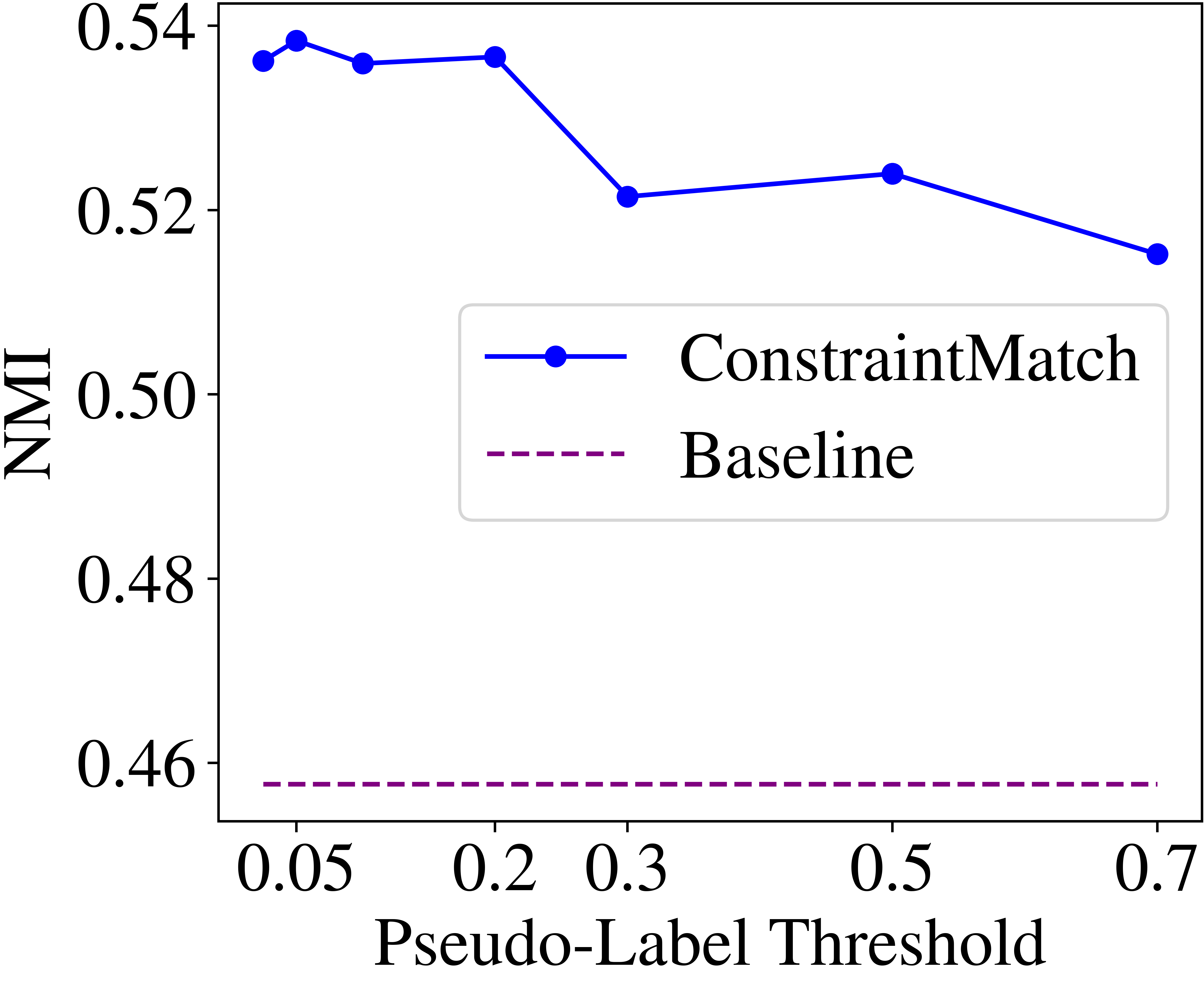}
        \caption{$\mathcal{I}^{sel}$: impact of $\tau$.}
        \vspace{0.35cm}
        \label{fig:analysis3}
    \end{subfigure}
    \begin{subfigure}[b]{0.3\textwidth}
        \captionsetup{width=.7\linewidth, size=scriptsize}
        \includegraphics[width=\textwidth]{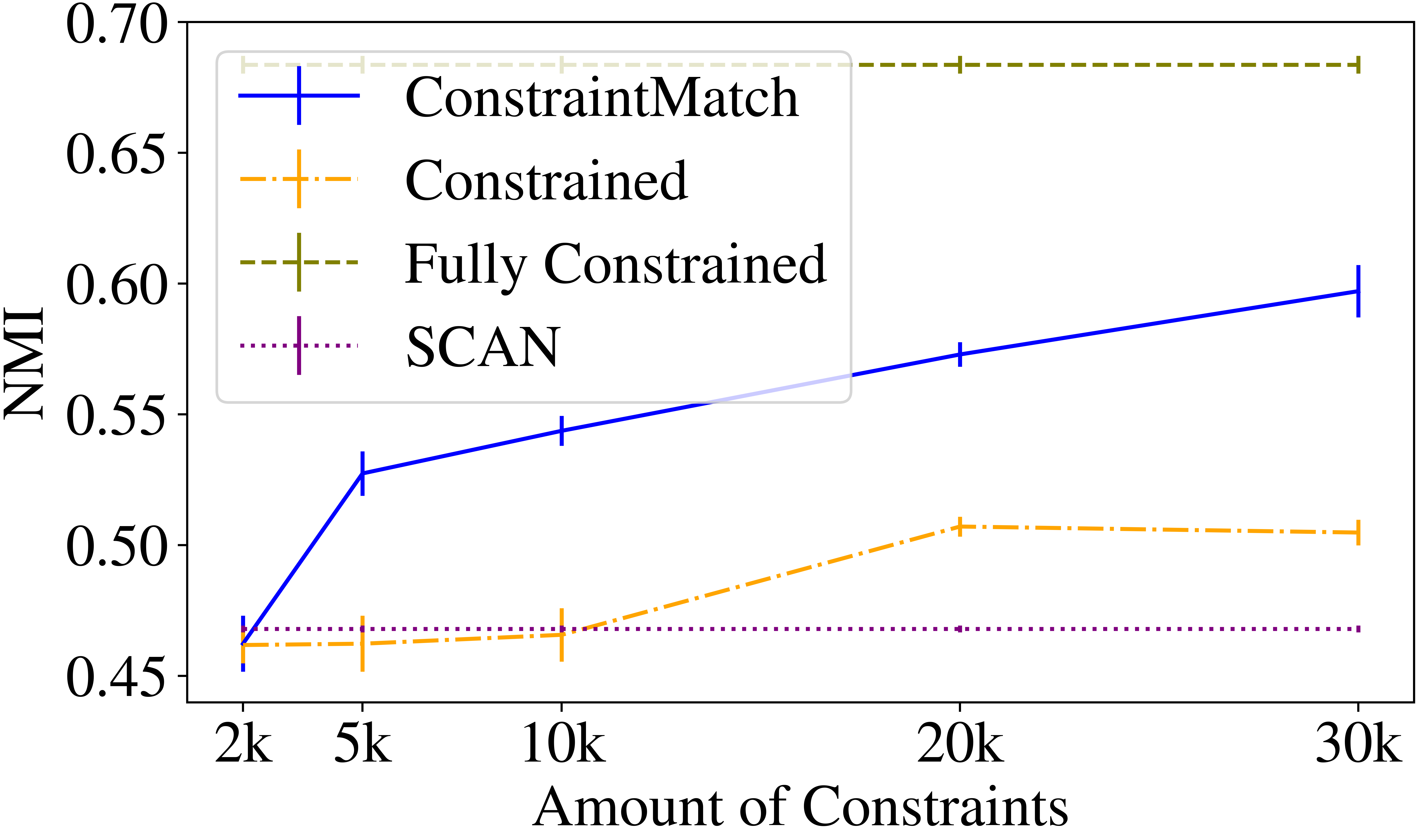}
        \caption{Effect of the amount of constraints $n_c$.}
        \label{fig:analysis4}
    \end{subfigure}
    \caption{Further analyses of ConstraintMatch.}
    \label{fig:analyses}
\vspace{-0.5cm}
\end{figure*}

In this section, we analyze the several components of ConstraintMatch. 
Unless noted otherwise, these analyses and experiments were run on the Cifar100-20 dataset with $n_c=10000$ using the optimal hyperparameters obtained for the main experiments and we report results on the test splits.

\textbf{Pseudo-Constraining}
\label{ch4:softconstraints}
\looseness=-1 We use soft pseudo-constraints $\tilde c_{ij} \in [0,1]$ in ConstraintMatch. 
One alternative would be the separation into hard pseudo-constraints $\tilde c_{ij}^h \in \{0, 1\}$ using a threshold $\mu$ such that $\hat c_{ij}^h = 1$ if $c_{ij} \geq \mu$ and $\hat c_{ij}^h = 0$ if $c_{ij} < \mu$. 
We find that while ConstraintMatch is still outperforming the constrained baseline using hard pseudo-constraints, it benefits further from the use of soft pseudo constraints as shown in Fig.~\ref{fig:analysis1} over different values of $\mu$.
Using soft constraints also eliminates the need to tune $\mu$.
We hypothesize that the model can effectively use the continuous information provided via the soft pseudo-constraints within the MCL loss, similar to the use of soft labels in supervised classification \cite{Meister2020}. 

\textbf{Pseudo-Label Selection}
\label{ch4:informativeness} 
\looseness=-1 We argue for the selection of \textit{informative} samples as pseudo-labels over that of \textit{confident} samples. 
Fig.~\ref{fig:analysis2} contrasts the use of both with a fixed value $\tau=0.2$ for the informativeness criterion and with varying thresholds for confidence-based selection showing that informative samples enable ConstraintMatch to leverage unconstrained samples more effectively.
Further, we provide a sensitivity analysis of the threshold $\tau$ within $\mathcal{I}^{sel}_{\tau}$ in Fig.~\ref{fig:analysis3}.
This reveals that the sensitivity of ConstraintMatch towards $\tau$ lies within a reasonable margin and we recommend $\tau \in [0.025, 0.2]$ as a range for tuning.

\textbf{Amount of Constraints}
\looseness=-1 Fig.~\ref{fig:analysis4} shows the effect of the amount of constraints $n_c$ on the model performance as measured in NMI over five folds. 
The constrained clustering competitor model performs worse than the SCAN baseline for $n_c \leq 10000$ which might be due to the fact that $n_c=10000$ results in $500$ Must-Link constraints only in the Cifar100-20 scenario, allowing the MCL loss to overfit those few pairs quickly.
In contrast to that, ConstraintMatch successfully overcomes this issue via its pseudo-constraining mechanism for $n_c \geq 5000$ with relative gains increasing for increasing $n_c$.
The comparably low performance of ConstraintMatch for $n_c=2000$ indicates that it still requires a certain degree of supervision to produce reliable pseudo-constraints.

\begin{figure}[h]
    \centering
    \includegraphics[width=6.5cm]{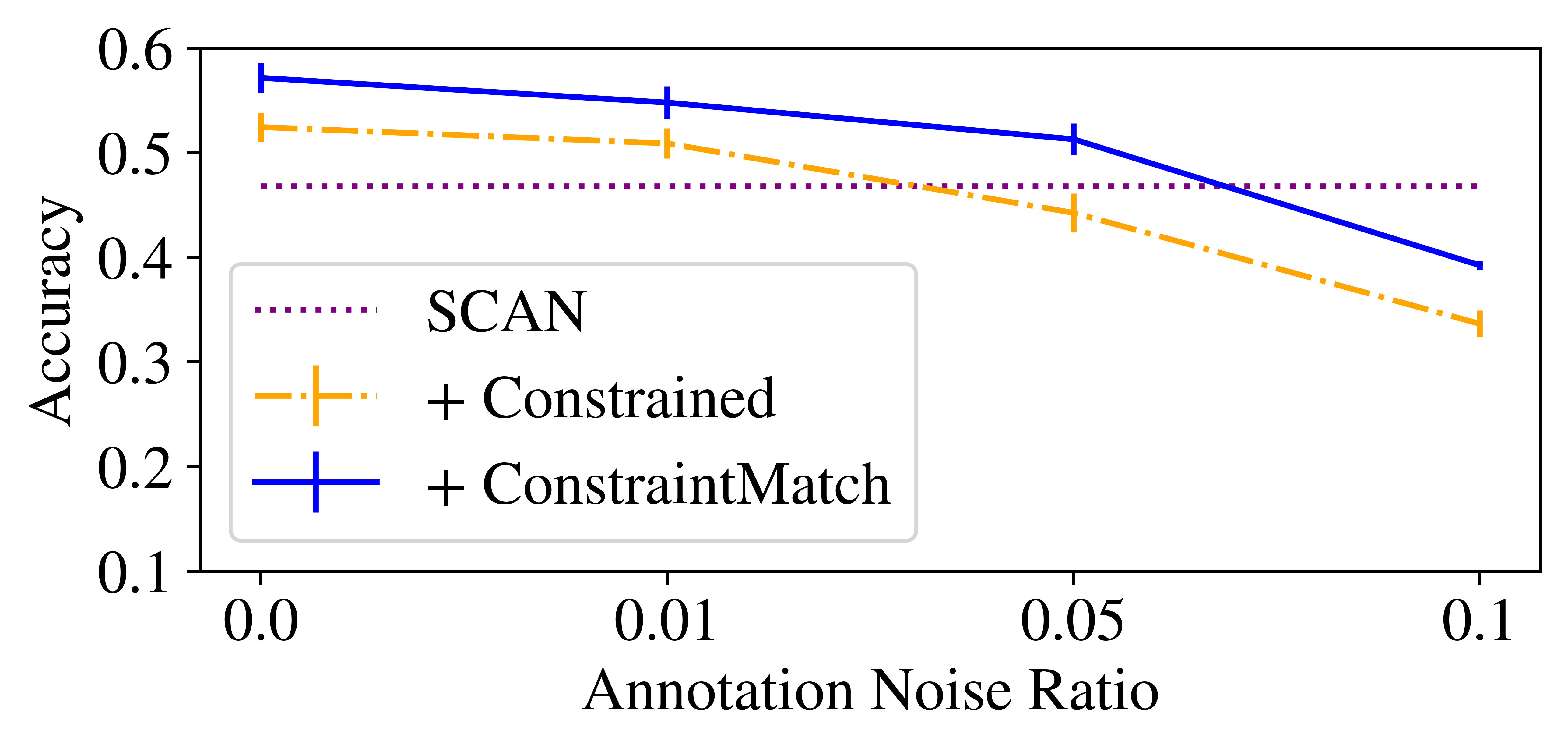}
    \caption{Impact of ground truth constraint annotation noise.}
    \label{fig:noisy_constraints}
\end{figure}

\textbf{Robustness w.r.t Noisy Constraints}
\looseness=-1 Next to the robustness of ConstraintMatch over noisy pseudo-labels, we further investigate its robustness towards noise in the annotation of the known ground truth constraints.
This simulates the situation where the annotators might erroneously flip the constraint annotation, similar to the concept of label noise in supervised classification \cite{Hedderich2021}. 
Therefore, we randomly flipped a varying percentage of the known constraint annotations and compared the effect of this annotation noise on model training. 
The results in Fig.~\ref{fig:noisy_constraints} show that ConstraintMatch is more robust towards higher levels of annotation noise than the constrained competitor. 
We attribute this increased robustness to the stabilizing effect of the pseudo-constraining mechanism.

\subsection{Overclustering}
\label{ch4:overcluster}

We further evaluate ConstraintMatch for overclustering, where the true amount of clusters $K$ is unknown and the model can assign more clusters than inherently present in the data, $n_{out}\gg K$ \cite{Hsu2016}.
Therefore, we compare ConstraintMatch with the constrained competitor and the SCAN baseline with $n_{out}=5K$ resulting in $100$ potential clusters for Cifar100-20 and $50$ for Cifar10.
Models were evaluated using the Hungarian Assignment \cite{Kuhn1955} with cluster predictions that do not match a corresponding ground truth cluster counting as an error. 
As shown in Table~\ref{tab:overclustering}, we find that the constrained competitor achieves strong performance gains over the unsupervised baseline despite the challenging task. 
ConstraintMatch's performance gains translate well to this overclustering scenario yielding a relative (absolute) performance gain over the constrained competitor of $18.11\%$ $(8.03\text{pp})$ NMI and $10.75\%$ $(4.24\text{pp})$ Accuracy on Cifar100-20. 

\begin{table}
    \caption{Overclustering results averaged over five folds.}
    \label{tab:overclustering}
    \centering
    \begin{tabular}{l c c c c c c}
        \toprule
        Dataset & \multicolumn{3}{c}{Cifar-10} & \multicolumn{3}{c}{Cifar 100-20}\\
        $n_{out}$ & \multicolumn{3}{c}{50} & \multicolumn{3}{c}{100}  \\
        \cmidrule(l{4pt}r{6pt}){2-4} \cmidrule(l{2pt}r{2pt}){5-7}
        Metric (Test) & ACC & NMI & ARI & ACC & NMI & ARI\\ 
        \midrule
        SCAN & 34.68 & 61.56 & 34.52 & 29.88 & 47.35 & 23.23 \\
        Constrained & 82.24 & 75.40 & 73.80 & 39.41 & 44.34 & 27.86 \\
        ConstraintMatch & \textbf{88.89} & \textbf{83.04} & \textbf{82.06} & \textbf{43.65} & \textbf{52.37} & \textbf{34.88} \\
        \bottomrule
    \end{tabular}
\end{table}

%
%

\section{Conclusion}

ConstraintMatch is a novel method for training clustering models in a semi-constrained setting, using a combination of large amounts of unconstrained data and a limited number of constraint pairs. 
Therefore, it selects informative pseudo-labels processed within a pseudo-constraining mechanism that allows training the model on a unified loss function to overcome the limitations of naive pseudo-labeling in this setting. 
With empirical results across five benchmarks, we demonstrate ConstraintMatch's strong performance, outperforming baselines and competitors by substantial margins, even in challenging overclustering scenarios. 
We furthermore support our algorithmic choices with empirical evidence and empirically showed that pseudo-constraining leads to increased model robustness towards different sources of annotation noise.

While we have shown the application in scenarios with a cluster cardinality of up to $K\leq20$, we did not yet investigate the applicability to high-cardinality settings.
Also, the use of alternative constraints such as triplets or continuous constraints as well as experiments with non-uniform annotation noise would be interesting for future research.

\section*{Acknowledgments}

JG and BB were partially supported by the Bavarian Ministry of Economic Affairs, Regional Development, and Energy through the Center for Analytics – Data – Applications (ADA-Center) within the framework of BAYERN DIGITAL II (20-3410-2-9-8) and the German Federal Ministry of Education and Research (BMBF) under Grant No. 01IS18036A, Munich Center for Machine Learning (MCML). 
ZK was partly supported by DARPA’s Learning with Less Labels (LwLL) program under agreement HR0011-18-S-0044


%
%

\newpage
\appendix

\subsection{Hyperparameter Tuning}
\label{app:tuning}

We conducted a grid search over the validation splits detailed in Table~\ref{tab:datalarge} for one fold of sampled training constraints for hyperparameter tuning. 
We used the validation loss on the constraints from the validation splits to select the optimal hyperparameters for each dataset and model combination with the lowest final validation loss as performance criterion. 
Final models were then trained on these optimal hyperparameters on five repeated folds of the respective constrained and unconstrained training samples and final performance metrics were reported for both the "Test" and the "Train(+Test)" settings.
The \textit{shared} parameters were used in and tuned for all trained models and the specific hyperparameters for ConstraintMatch and the naive pseudo-labeling baseline were tuned over a grid of different values, see Table~\ref{tab:hpars}.

\begin{table}[h]
\caption{Hyperparameters and their respective values considered in the grid search for the different models.}
\centering
\begin{tabular}{ll}
    \toprule
    Parameter & Search Values\\
    \midrule
    \multicolumn{2}{c}{Shared}\\
    \midrule
    Weight decay  & $0.001$, $0.0001$, $0.00001$\\
    Learning rate & $0.03, 0.01, 0.003, 0.001, 0.0001$\\
    \midrule
    \multicolumn{2}{c}{naive Pseudo-labeling}\\
    \midrule
    $\lambda$ & $1.0, 0.5, 0.1, 0.05$ \\
    $\tau$ & $0.7, 0.8, 0.9, 0.95, 0.99$ \\    
    \midrule
    \multicolumn{2}{c}{ConstraintMatch}\\
    \midrule
    $\lambda$ & $1.0, 0.5, 0.1, 0.05$ \\
    $\tau$ & $0.05, 0.1, 0.2, 0.3$\\
    \bottomrule
\end{tabular}
\label{tab:hpars}
\end{table}

\subsection{Data Augmentation}
\label{app:augmentation}

ConstraintMatch follows the rationale of consistency regularization via weak and strong augmentations $a()$ and $A()$.
As weak augmentations $a()$, we used random cropping and horizontal flipping. 
For strong augmentations, $A()$, we used the RandAugment strategy with the data augmentation procedures used in FixMatch and described in Appendix D of \cite{Sohn2020}. 

\subsection{Datasets}
\label{app:data}

Table~\ref{tab:datalarge} provides an overview of the datasets used in the experimental section~\ref{ch4:experimentalsetup} alongside their splits and sizes.
The final column describes the exact dataset splits that were used in the Train(+Test) evaluation setting following \cite{Li2021}.

\begin{table}[h!]
    \caption{Datasets used in the experiments including the respective training, validation, and test splits. We also mention the evaluation dataset for the Train(+Test) setting in the last column following \cite{Li2021}.}
    \centering
    \begin{tabular}{l c c c c}
        \toprule
        Dataset & $K$ & Samples & Constraints & Train(+Test)\\
        & & Train Val Test & Train Val &\\
        \midrule
        Cifar10&10& 45k 5k 10k & 5/10k 10k & Train + Test\\ 
        Cifar100-20& 20 & 45k 5k 10k & 5/10k 10k & Train + Test \\ 
        STL10& 10 & 4k 1k 8k & 5/10k 5k & Train + Test\\ 
        ImageNet-10& 10 & 12k 1k 500 & 5/10k 1k & Train \\ 
        ImageNet-Dogs& 15 &18.5k 1k 750 & 5/10k 1k & Train \\
        \bottomrule
    \end{tabular}
    \label{tab:datalarge}
\end{table}

\subsection{Visualization of the Confirmation Bias}

\begin{figure}
    \centering
    \includegraphics[width=0.3\textwidth]{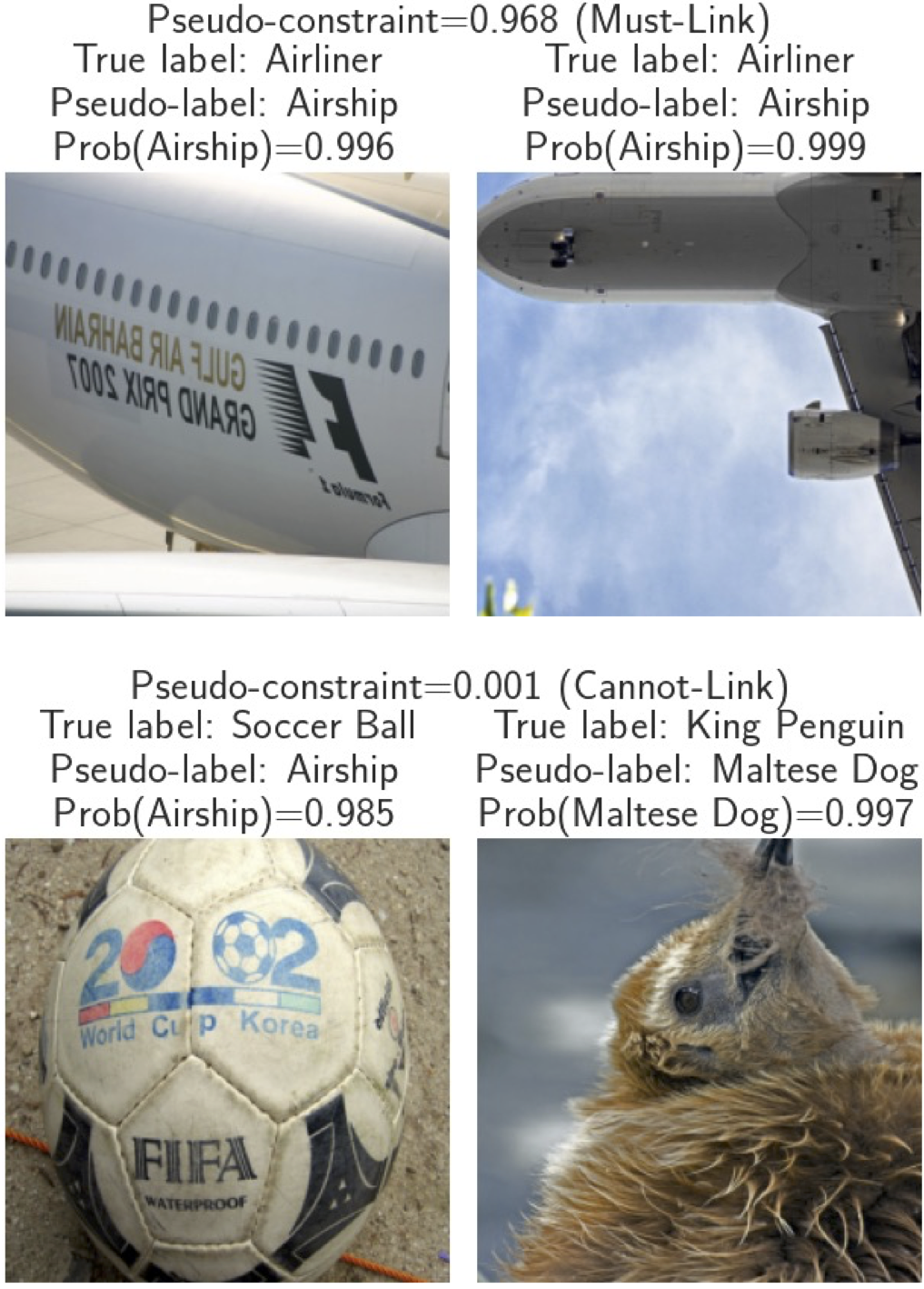}
    \caption{
        Illustration of pseudo-labeling failure cases due to confirmation bias. 
        Pseudo-constraints generated on top of these wrong pseudo-labels are still semantically correct.
    }
    \label{fig:failed_pls}
\end{figure}

In Fig.~\ref{fig:failed_pls}, we visualized four samples from the unconstrained part of the ImageNet-10 dataset which suffer from the confirmation bias similar to Fig.~\ref{fig:pc_motivation}, i.e. samples for which the model confidently predicted the wrong cluster assignment.
These unconstrained samples were selected as high-confidence (max. predicted probability $>0.98$) but wrongly predicted examples. 
We can observe that pseudo-labeling would lead to wrong prediction targets (e.g. cluster "Airship" instead of the true cluster "Soccer Ball" in the bottom left example) and hence confuse model training. 
On the other hand, pseudo-constraints generated on top of pairs of these wrongly assigned pseudo-labels still are semantically correct and can support model training on these unconstrained samples. 
This does not only hold for Cannot-Link (bottom) but also for Must-Link (top) pseudo-constraints where both samples with the same true cluster affiliation are assigned the same wrong cluster by the model. 
These samples were selected from a random batch of unconstrained samples $\mathcal{B}^u$ from the ImageNet-10 dataset from ConstraintMatch trained for 500 training steps with a ResNet-34 backbone.

\subsection{Naive Pseudo-Labeling Baseline}
\label{app:npl}

\begin{figure*}
    \centering
    \begin{subfigure}[b]{\textwidth}
        \centering
        \includegraphics[width=12cm]{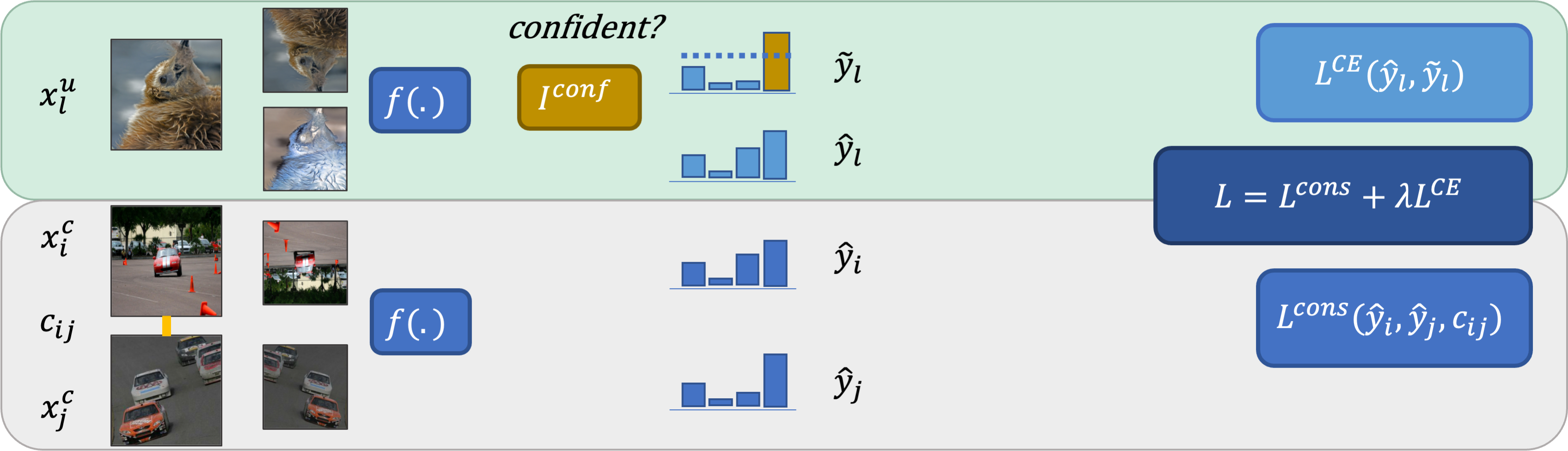}
        \caption{The naive pseudo-labeling baseline combines training on pairwise constrained (gray) and individual unconstrained (green) samples leveraging weak and strong data augmentations following the FixMatch approach \cite{Sohn2020}. 
        A cross-entropy loss function is used as an auxiliary loss function for the pseudo-labeled unlabeled samples.}
        \label{fig:baseline_npl}
    \end{subfigure}
    \begin{subfigure}[b]{\textwidth}
        \centering
        \vspace{0.5cm}
        \includegraphics[width=12cm]{ccm_large.png}
        \caption{ConstraintMatch combines pairwise training on constrained (gray) and unconstrained (yellow) samples leveraging weak and strong data augmentations. It extends the naive pseudo-labeling baseline by the generation of pseudo-constraints from \textit{informative} pseudo-labels to overcome the confirmation bias as detailed in the methods section of the paper.}
    \label{fig:baseline_constraintmatch}
    \end{subfigure}
    \caption{Illustration of a) the naive pseudo-labeling baseline and b) ConstraintMatch.}
    \label{fig:baseline_npl_cm}
\vspace{-0.5cm}
\end{figure*}

In Fig.~\ref{fig:baseline_npl_cm}, we visualize the naive pseudo-labeling baseline, a simplified version of ConstraintMatch, with which we compared ConstraintMatch in the results Section~\ref{ch4:results}. 
This baseline follows the use of unlabeled samples in FixMatch \cite{Sohn2020} and similarly leverages the weak-strong augmentation scheme for consistency regularization. 
Concretely, weakly augmented, unconstrained samples are selected via a confidence threshold over their predicted cluster assignments, and the predicted clusters with the highest assigned probability are subsequently chosen as pseudo-labels.
This confidence-based selection criterion is depicted as $\mathcal{I}^{conf}$ in Fig.~\ref{fig:baseline_npl} and the associated threshold $\tau \in [0, 1]$ is a hyperparameter that we tuned on the validation set as described above and listed in Table~\ref{tab:hpars}.
Predictions over strong augmented versions of these samples then serve as input for an auxiliary cross-entropy loss function, referred to as $\mathcal{L}^{CE}$ in Fig.~\ref{fig:baseline_npl}. 
Similar to ConstraintMatch, the constrained loss $\mathcal{L}^{cons}$ is calculated over model predictions on pairwise samples and their corresponding constraint annotations. 
The naive pseudo-labeling baseline is then trained via the final loss $\mathcal{L}=\mathcal{L}^{cons} + \lambda \mathcal{L}^{CE}$ as a weighted linear combination of both losses where hyperparameter $\lambda$ controls the impact of the unconstrained samples.

\end{document}